\documentclass{article}

\usepackage{arxiv}
\usepackage{amsmath}
\usepackage{enumitem}

\usepackage{latexsym}
\usepackage{makecell}

\usepackage[T1]{fontenc}
\usepackage[utf8]{inputenc}

\usepackage{booktabs}

\usepackage{microtype}

\usepackage{graphicx}

\usepackage{listings}
\usepackage{xcolor}
\usepackage[round,authoryear]{natbib}
\let\cite\citep
\usepackage{hyperref}

\title{%
  \raisebox{-0.35ex}{\includegraphics[width=0.8cm]{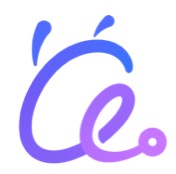}}%
  \hspace{3.5px}\emph{MedResearcher-R1}: Expert-Level Medical Deep Researcher via A Knowledge-Informed Trajectory Synthesis Framework%
}

\author{
Ailing Yu$^{1}$ \quad
HaoYang Hong $^{2}$ \thanks{Work done during an internship at Ant Group.} \quad
Lan Yao$^{1}$ \quad
Jingnan Liu$^{1}$ \quad
Zhe Chen$^{1}$ \quad
Jiajun Yin$^{1}$ \\
Yuan Wang$^{1}$ \quad
Xinhao Liao$^{1}$ \quad
Zhiling Ye$^{1}$ \quad
Ji Li$^{1}$ \quad
Yun Yue$^{1}$ \\
Hansong Xiao$^{1}$ \quad
Hualei Zhou$^{1}$ \quad
Chunxiao Guo$^{1}$ \quad
Peng Wei$^{1}$ \quad
Junwei Liu$^{1}$ \quad
Jinjie Gu$^{1}$ \\[1ex]
$^{1}$Ant Group \qquad
$^{2}$Imperial College London
}

\date{}
\renewcommand{\shorttitle}{MedResearcher-R1}

\begin{document}
\maketitle

\begin{center}
\begin{abstractbox}
% Recent advances in Large Language Model (LLM)-based agents have enabled strong performance in deep research tasks, yet they remain limited in the medical domain.
Large Language Model (LLM)-based agents have recently demonstrated remarkable capabilities in deep research tasks, yet their application to the medical domain remains challenging. The leading proprietary systems achieve only modest results on complex medical benchmarks. This exposes two key bottlenecks: (i) limited clinical-reasoning capability in the underlying model, and (ii) unreliable access to authoritative medical evidence when constrained to open-web retrieval. To address these limitations and equip agents with robust clinical reasoning capabilities, we present MedResearcher-R1, a medical deep research agent featuring two core innovations.
% To empower the agent with real clinical reasoning ability, we introduce MedResearcher-R1, a medical deep research agent that addresses these challenges with two key innovations. 
First, we propose Knowledge-Informed Trajectory Synthesis (KISA), a novel approach that builds medical knowledge graphs to construct complex multi-hop question–answer pairs centered on rare medical entities, overcoming the scarcity of high-quality domain-specific training data.
Second, we integrate a medical retrieval engine alongside general-purpose tools, enabling precise and reliable synthesis of medical evidence.
Through this methodology, we yield over 2,100 diverse trajectories spanning 12 medical specialties. 
Trained with supervised fine-tuning and reinforcement learning, our MedResearcher-R1-32B achieves state-of-the-art performance on MedBrowseComp (27.5/50 vs. 25.5/50 for o3-deepresearch) while demonstrating strong general performance on GAIA and xBench benchmarks. 
To the best of our knowledge, we present the first high-quality, tool-augmented medical dataset paired with a domain-specialized deep-research agent, demonstrating that smaller open-source models can surpass substantially larger proprietary systems on specialized medical tasks.
\end{abstractbox}
\end{center}

\section{Introduction}
Recent advances in Large Language Models (LLMs) have catalyzed the emergence of deep research systems. Unlike static question answering, deep research systems maintain evolving hypotheses, actively acquire external evidence via search and specialized tools, reconcile conflicting information, and synthesize grounded conclusions into structured artifacts (e.g., reports) over multi-step trajectories~\citep{xu2025comprehensivesurveydeepresearch}. 
Despite rapid progress, the medical domain imposes stricter requirements on reasoning validity and evidence authority, which current general-purpose deep research agents still fail to satisfy reliably~\citep{chen2025medbrowsecompbenchmarkingmedicaldeep}.

First, the reasoning architectures of general agents are often too rigid for the stochastic nature of medical inquiry. Standard agent workflows typically follow linear "plan-and-execute" trajectories. However, medical research is inherently abductive and iterative. It requires agents to constantly re-evaluate hypotheses and cross-validate conflicting data points~\citep{yu2025augmenting}. 

Second, there is a fundamental disconnect between general pre-training and clinical data. General deep research agents are trained to optimize for plausible text generation rather than faithful clinical reasoning. They lack exposure to the specific cognitive trajectories used by medical professionals~\citep{wang2025medicalreasoningerallms}. While Reinforcement Learning (RL) offers a potential solution, it often struggles in this domain due to the sparsity of intermediate reasoning signals. Consequently, Supervised Fine-Tuning (SFT) on expert-demonstrated trajectories emerges as the most trustful strategy. However, existing data synthesis pipelines for SFT remain insufficient. They mainly focus on generating static, text-only samples~\citep{quan2024automaticallygeneratingnumerouscontextdriven, nadas2025syntheticdatagenerationusing} or unifying disparate benchmarks into standardized schemas~\citep{song2025agentdataprotocolunifying}.

Third, general agents suffer from a bottleneck due to inadequate information retrieval. Most deep research systems rely on commercial search engines designed for consumer relevance rather than scientific accuracy. In medicine, where the "ground truth" is often sequestered in specialized, authoritative indices (e.g., FDA databases), reliance on open-web retrieval leads to the non-authoritative or outdated information~\citep{lin2025llmbasedagentssufferhallucinations}. This "retrieval gap" is a primary driver of hallucination in medical agents, as they attempt to bridge missing evidence with plausible but fabricated details.

\begin{figure}[t]
    \centering
    \includegraphics[width=0.975\linewidth]{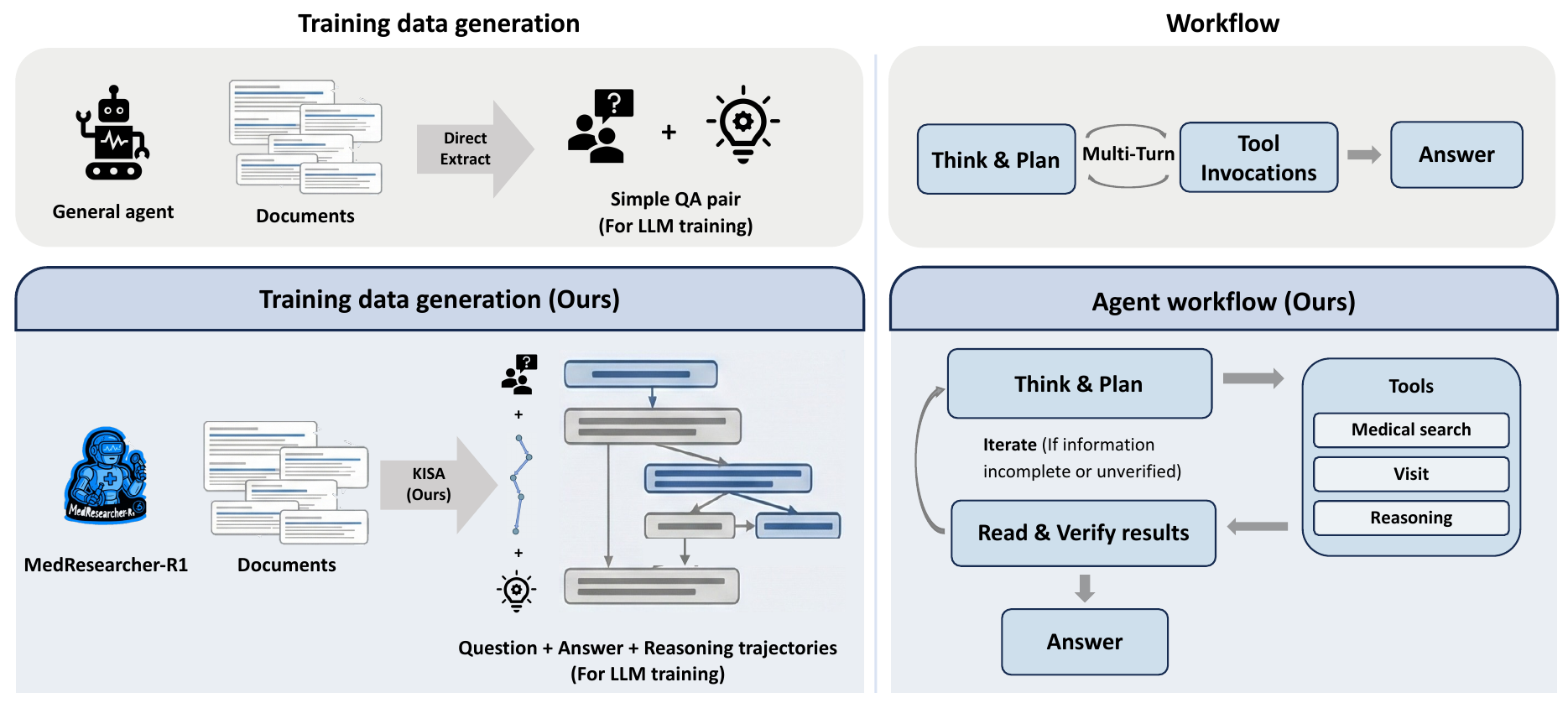}
    \caption{\textbf{Comparison of medical reasoning agents}. MedResearcher-R1 improves general-purpose agents from both the training-data and architectural perspectives. On the training data side, it is trained on a specially designed medical dataset and guided by reasoning trajectories. On the architectural side, it employs a multi-turn tool-invocation workflow based on a think, tool invocation, verification loop and is equipped with medical-specific tools.}
    \label{fig:agent}
\end{figure}

To bridge these gaps, we propose a framework that redefines how medical agents are designed and trained. We posit that robust medical reasoning demands both a domain-adaptive workflow and exposure to complex, knowledge-rich scenarios during training. Our contributions are as follows:
\begin{itemize}[leftmargin=12.5pt, itemsep=-0.2em]
    \item \textbf{Domain-adaptive workflow for medical deep research.} We introduce a multi-turn, iterative workflow (Figure \ref{fig:agent}) where the agent alternates between internal reasoning ("think") and external tool invocation. A central feature of this architecture is the cross-validation of retrieved evidence to ensure clinical accuracy. The clinical accuracy is critical requirement in medicine, where reliance on partial information often leads to hallucination or error.
    \item \textbf{Knowledge-Informed Trajectory Synthesis (KISA).} We propose KISA, a novel data generation framework for supervised fine-tuning deep research agents. KISA leverages biomedical knowledge graphs to synthesize high-quality reasoning trajectories, effectively teaching the agent how to navigate the problem space and utilize tools strategically for deep research tasks.
    \item \textbf{Comprehensive medical tool integration.} We augment the agent's capabilities with a custom retrieval engine accessing authoritative data sources. Unlike general agents, our system directly queries FDA databases, official prescription records, clinical trial registries, and peer-reviewed literature to ground its reasoning in verified medical data.
\end{itemize}

% \vspace{-0.2em}
% Our evaluation spans both medical and general-purpose deep research settings:
% (1) MedResearcher-R1 achieves state-of-the-art medical agent performance with \textbf{27.5/50} tasks solved on MedBrowseComp as is shown in Table~\ref{tab:med-results}, surpassing o3-deepresearch (25.5/50). 
% (2) Beyond the medical setting, MedResearcher-R1 exhibits strong general capability, scoring \textbf{53.4} on GAIA~\citep{gao2023gaia} and \textbf{54.0} on XBench~\citep{chen2025xbench} (Table~\ref{tab:main-results}). 
% These results challenge the assumption that domain specialization requires sacrificing general capabilities, demonstrating that medical training can enhance rather than limit the versatility of agents.

\section{Related Work}
% This field is rapidly advancing primarily through two paradigms. One line of work focuses on multi-agent planning architectures, where complex tasks are decomposed and assigned to specialized agents for retrieval, reasoning, and synthesis~\citep{li2023camel, xu2025comprehensive}. Another prominent approach uses reinforcement learning (RL) to train a single agent to interact with complex web environments, optimizing its reasoning and tool-use strategies through reward signals~\citep{yao2023reactsynergizingreasoningacting, li2025websailor}. These systems, exemplified by state-of-the-art models like WebSailor and Search-R1, have demonstrated superhuman performance on general web research tasks.

\textbf{LLM-based Agents for Deep Research}. Deep research (DR) systems equip LLMs to generate evidence-rich reports. Optimization strategies generally fall into three categories: workflow engineering, supervised fine-tuning, and reinforcement learning (RL)~\citep{shi2025deepresearchsystematicsurvey}. The RL paradigm has yielded agents with great performance such as Search-R1~\citep{jin2025searchr1trainingllmsreason}, DeepResearcher~\citep{zheng2025deepresearcherscalingdeepresearch}, WebDancer~\citep{wu2025webdancerautonomousinformationseeking}, and Kimi-K2~\citep{kimiteam2025kimik2openagentic}, while workflow engineering is exemplified by systems like Gemini DeepResearch~\citep{geminideepresearch2024}, OpenAI DeepResearch~\citep{openaideepresearch2025}, Manus~\citep{manusdeepresearch}, and SunaAI~\citep{sunaaideepresearch}.

Despite these advances, performance degrades in high-stakes fields like medicine~\citep{chen2025medbrowsecompbenchmarkingmedicaldeep}. General-purpose agents often lack the specialized retrieval tools and reasoning depth necessary for clinical synthesis. Consequently, they struggle to navigate authoritative medical databases or capture relationships between rare entities, a limitation this work addresses.

\noindent
\textbf{Deep research agents for medicine}. In the medical domain, AI methodologies have evolved from isolated diagnostic models to integrated agentic systems. Initial approaches primarily leveraged Retrieval-Augmented Generation (RAG) to ground clinical decision-making in external literature~\citep{zhao2025medragenhancingretrievalaugmentedgeneration, toma2023clinicalcamelopenexpertlevel}. Recently, this has expanded into multi-agent frameworks designed for specialized clinical workflows, such as sequential diagnosis~\citep{nori2025sequentialdiagnosislanguagemodels} and dynamic knowledge management~\citep{yao2024seakrselfawareknowledgeretrieval}.

However, current agents struggle with exploratory medical research. Performance notably deteriorates on tasks requiring deep, multi-step inference~\citep{schmidgall2025agentclinicmultimodalagentbenchmark}. This brittleness is attributed to a lack of training data which can reflect the high complexity of real-world medical investigation, such as linking rare diseases with disparate and authoritative findings. Therefore, developing high-quality training datasets for supervised fine-tuning medical DR system is necessary.

\noindent
\textbf{SFT training data for deep research agents}. Despite the success of RL and workflow engineering, SFT for DR agents is hindered by a lack of training data. Standard synthetic data methods for LLMs~\citep{quan2024automaticallygeneratingnumerouscontextdriven, nadas2025syntheticdatagenerationusing} are unsuitable for agentic systems. This is because they lack the requisite tool-use trajectories (e.g., <tool\_call>). While the Agent Data Protocol (ADP) successfully standardizes diverse agent datasets into a common schema~\citep{song2025agentdataprotocolunifying}, it focuses on organizing existing datasets rather than generating agentic training data from scratch. This can be particularly acute for medical DR agents training due to the lack of datasets in this domain. Existing approaches regarding SFT training data primarily focus on instructing LLMs to utilize functions and invoke real-world APIs \cite{patil2023gorillalargelanguagemodel, qin2023toolllmfacilitatinglargelanguage, tang2023toolalpacageneralizedtoollearning, chen2024enhancingfunctioncallingcapabilitiesllms}. However, such datasets fall short in enhancing the complex reasoning capabilities required for deep research. Consequently, we aim to propose a data construction method specifically designed to augment the cognitive processes of deep research agents.

\section{MedResearcher-R1: Medical Deep Research Agent Framework}

\label{sec:agent_architecture}
% \vspace{-0.3em}

To adapt the agent workflow for solving deep medical research problems, we propose a \textit{think-tool invocation-verification} loop (Figure \ref{fig:agent}). This process occurs before the final answer is generated. Specifically, given a query, MedResearcher-R1 performs the following steps: (i) \textbf{Think}: In this initial step, the agent plans the research trajectory, identifies required information, and determines which parts of the retrieved evidence require cross-validation to ensure accuracy. It also selects the appropriate tools and defines their specific tasks. (ii) \textbf{Tool Invocation}: The agent executes the tools selected during the "Think" phase. (iii) \textbf{Verification}: This step conducts consistency checks on the evidence flagged for cross-validation. It compares the evidence from different sources and draw a conclusion. Furthermore, once verification is complete and sufficient evidence has been accumulated, this module synthesizes the findings to generate the final evidence-based response.

The tool set aims to retrieve comprehensive medical information. To do this, we add a specialized Medical Search tool to standard utilities like web search and document reading. This specific tool queries authoritative medical repositories. Key sources include FDA datasets, prescription databases, clinical trial registries, and PubMed.

\section{KISA: Knowledge-informed trajectory synthesis approach}\label{sec:kisa}
\begin{figure}[t]
    \centering
    \includegraphics[width=\linewidth]{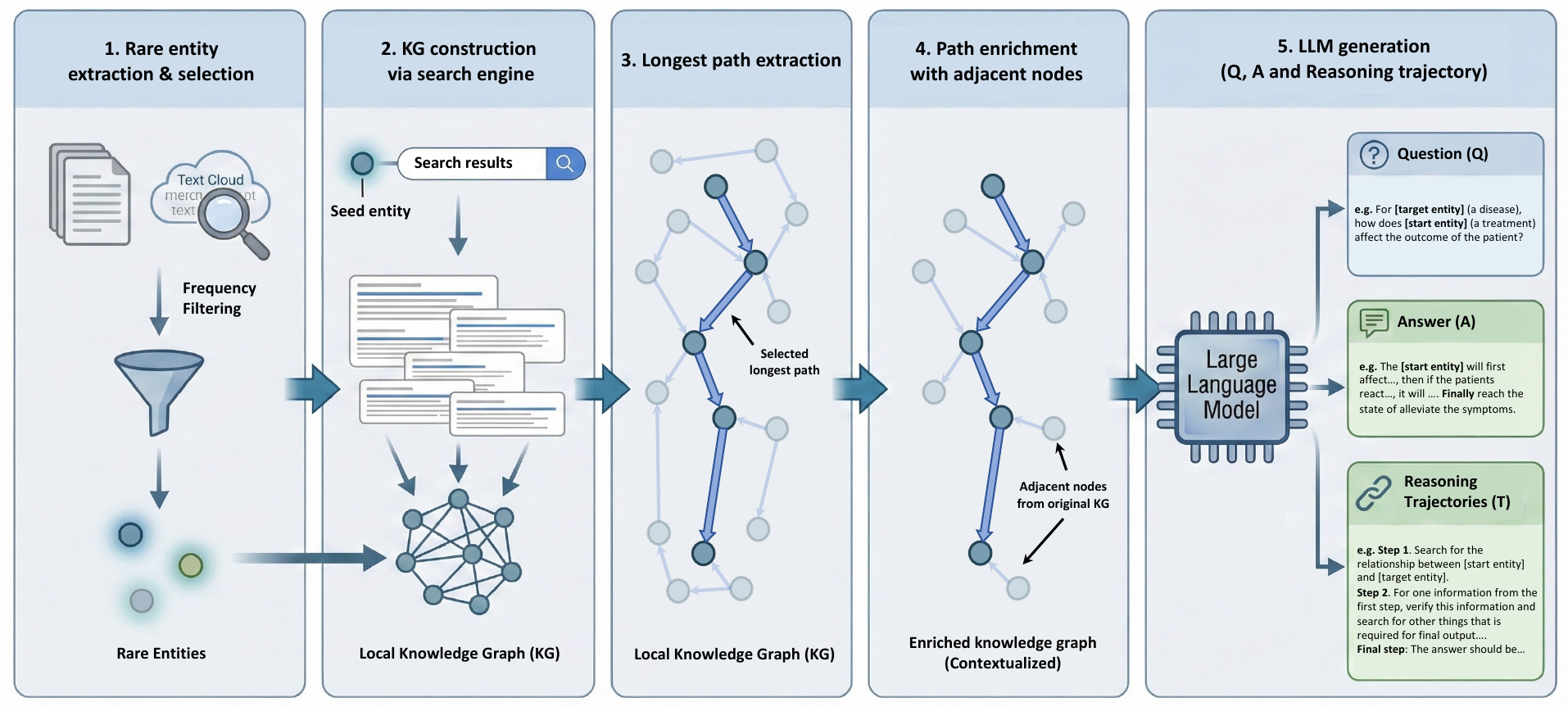}
    \caption{\textbf{KISA process illustration}. The process aims to generate explicit reasoning trajectories for agent SFT.}
    \label{fig:kisa_explanation}
\end{figure}

We want to use Supervised Fine-Tuning (SFT) to train MedResearcher-R1, which helps it to learn reasoning steps rather than training with reinforcement learning for random exploration. To address the critical challenge of training data scarcity for medical deep-research agents, we propose a Knowledge-Informed Trajectory Synthesis Approach (KISA) that generates complex, multi-hop medical reasoning trajectories for agent SFT. KISA consists of five stages (Figure \ref{fig:kisa_explanation}): (i) rare entity extraction and selection (ii) knowledge graph construction via search engine (iii) longest path extraction (iv) path enrichment with adjacent nodes and (v) reasoning trajectory generation with masked trajectory guidance. We now describe each of these stages in detail.

% \vspace{-0.5em}
% \subsection{Knowledge graph construction}

% \vspace{-0.3em}
A knowledge graph represents real-world entities and the relationships among them. This is particularly useful for medical reasoning because it encodes tractable relationships between medical concepts. Therefore, we aim to generate questions with reasoning trajectories for agent SFT from a medical knowledge graph. 

\noindent
\textbf{Rare entity extraction and selection}. Inspired by large-scale efforts to construct knowledge graphs from literature \citep{Kostis2020}, we begin by extracting medical terminologies as entities from over 30 million PubMed abstracts. Specifically, we use Qwen3-235B \cite{yang2025qwen3technicalreport} with carefully designed prompts to extract precise entities. This process also returns basic statistics such as frequency of entities, which helps screening. Additionally, the entity set for the knowledge graph is constructed incrementally. Each time we consider adding a new entity to the entity set, we require LLM to determine whether it should be added as a distinct entity based on current entity set. This procedure helps us handle typos and different representations referring to the same entity. After the entity set is constructed, we select rare entities whose frequencies are less then a threshold ($10^{-6}$).

\noindent
\textbf{Knowledge graph construction via search engine}. We focus on an entity $e \in \mathcal{E}$ that represents a rare medical term. For this entity, we construct a knowledge graph $\mathcal{G}(e) = (\mathcal{E}(e), \mathcal{R}(e), \mathcal{T}(e))$. $\mathcal{E}(e)$ denotes the entities, and $\mathcal{R}(e)$ denotes the relation types. The set $\mathcal{T}(e)$ contains triples in the form $(h,r,t)$. These represent a head $h$, a relation $r$, and a tail $t$. To construct the graph, we query the seed entity $e$ in a search engine. This retrieves data from both general and medical resources. We extract relational triples from the search results. These triples form the structure of $\mathcal{G}(e)$. 

\noindent
\textbf{Longest path extraction and enrichment}. We extract the longest path $P^*(\mathcal{G}(e))$ from $\mathcal{G}(e)$, i.e.
$$P^*(\mathcal{G}(e))=\arg\max_{P\in\mathcal{P}(\mathcal{G}(e))} |P|$$
where $\mathcal{P}(G(e))$ denotes the set of paths in graph $\mathcal{G}(e)$ and $|P|$ denotes the length (number of edges) of path $P$.
This strategy ensures that questions require multiple reasoning hops. Next, we retrieve the adjacent nodes for every node in this path. We use these to expand the path into a sub-knowledge graph $\mathcal{G}_{sub}(e)$. 
Let $V(P)$ denotes the set of nodes in path $P$ and $\mathcal{N}(e)$ denotes the neighbors of entity $e$ in graph $\mathcal{G}(e)$. Then
\vspace{0.25em}
\begin{align*}
    &\mathcal{G}_{sub}(e)=(\mathcal{E}_{sub}(e), \mathcal{R}_{sub}(e), \mathcal{T}_{sub}(e))\\
    &\mathcal{E}_{sub}(e)=\cup_{v\in V(P^*(\mathcal{G}(e)))}\mathcal{N}(v)
\end{align*}
\vspace{0.25em}
This provides extra information to accurately describe the nodes. This context supports the masked trajectory guidance method. Finally, we use $\mathcal{G}_{sub}(e)$ to generate question and answer pairs with reasoning trajectories.

\noindent
\textbf{Reasoning trajectory generation with masked trajectory guidance (MTG)}.
We input the sub-knowledge graph into LLMs. They generate questions, answers, and reasoning trajectories in natural language. Domain experts verify the extracted paths to ensure medical relevance. Additionally, we enforce adaptive difficulty calibration. We test each question with OpenAI-o3 and GPT-4. If either model achieves accuracy greater than $50\%$, the system regenerates the question. This ensures sufficient complexity.

\begin{figure}[t]
    \centering
    \includegraphics[width=0.915\linewidth]{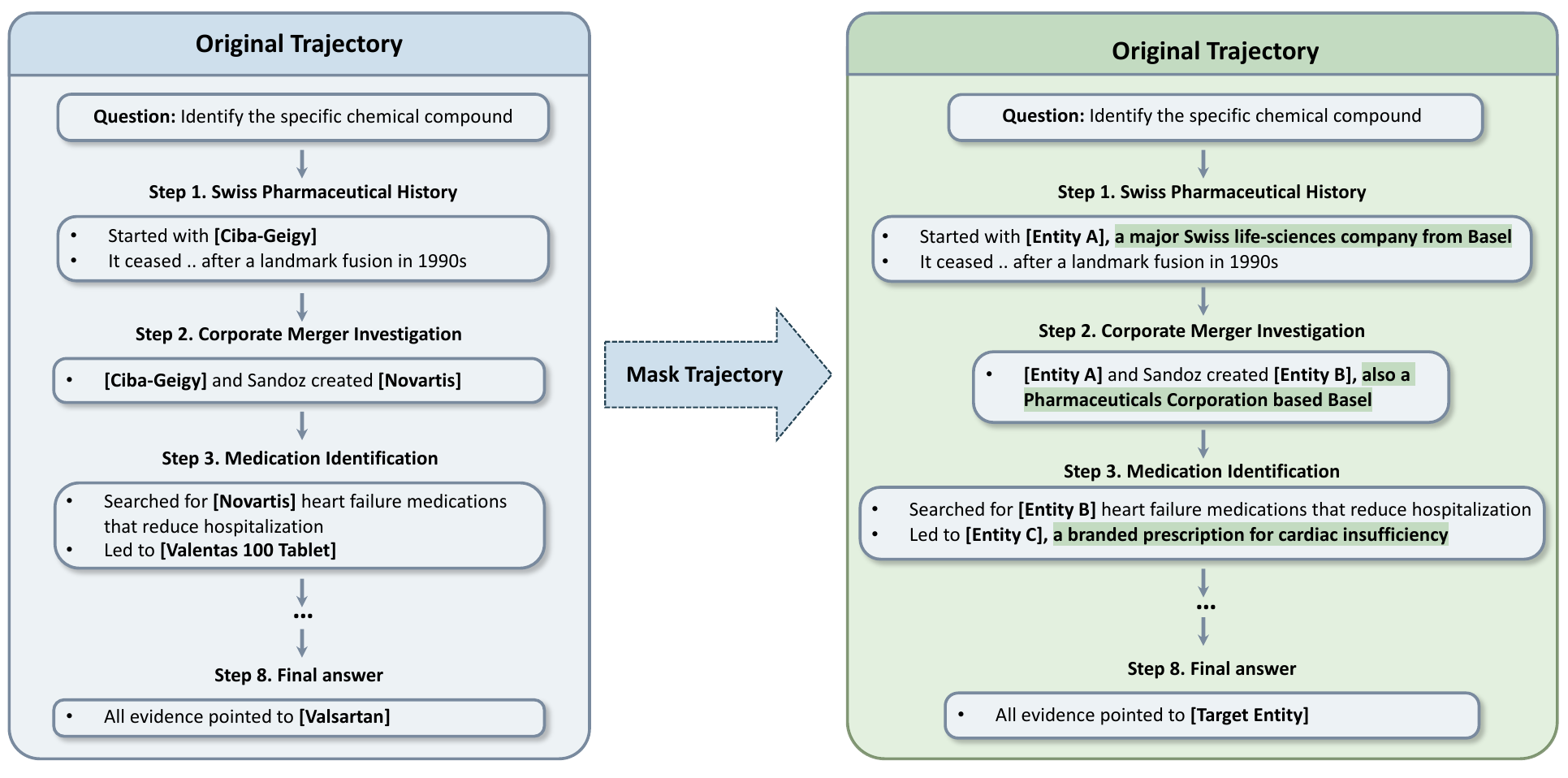}
    \caption{\textbf{Masked trajectory guidance sample}. The entities in the original reasoning path are masked with blurred descriptions. \textbf{[bold]} notation indicates entities from the knowledge graph on which this reasoning trajectory is constructed. The green-shaded texts are descriptions from the additional nodes.}
    \label{fig:masked_hint}
\end{figure}

This process yields Q\&A pairs with reasoning trajectories. Figure \ref{fig:masked_hint} (left) illustrates an example. The reasoning trajectories describe how to solve the given question from the initial entity step by step. However, these text trajectories lack tool-call indicators. They also lack observations from tool invocations which are necessary for agent training. Therefore, we propose Masked Trajectory Guidance (MTG). This method generates comprehensive reasoning paths for agent SFT. It incorporates the thinking process, tool tokens, invocation observations, and the final answer.

\textbf{Masked trajectory guidance (MTG)}. We employ a powerful LLM to generate trajectories for supervised fine-tuning. Our goal is to guide the reasoning process. However, we must not leak information that requires retrieval from tool invocations. To achieve this, we input the natural language trajectories alongside the question. We mask key entities within these trajectories. Specifically, MTG identifies entities present in the extracted path $P^*(\mathcal{G}(e))$. It replaces them with placeholders, such as "Entity A". It also adds context descriptions based on adjacent nodes in the sub-knowledge graph. Figure \ref{fig:masked_hint} illustrates an example. We feed the questions with the masked guidance to OpenAI-o3 \cite{openaio3systemcard2025}, which is equipped with our tool packages. The model generates complete reasoning trajectories, including the thinking process, tool tokens, and final answers. This output is ideal for agent SFT. We store the resulting data in $\mathcal{D}_{guided}$.

We also control the data mix. This prevents over-reliance on trajectory guidance in case there are easier reasoning ways to reach the answers. In specific, we create another training dataset called $\mathcal{D}_{explore}$. Here, we input only the questions into OpenAI-o3. We do not provide masked trajectories for detailed guidance. If the LLM generates a correct answer, we accept the associated trajectory. This ensures the model learns tool invocations naturally. It also prevents overfitting to the longest reasoning paths.

The final training dataset can be denoted as $\mathcal{D}_{train}=\{(x_i, y_i)\}_{i=1}^N$. Here, $x_i$ is the medical question. $y_i$ contains the answer and the reasoning trajectory. This dataset combines $\mathcal{D}_{guided}$ and $\mathcal{D}_{explore}$. In $\mathcal{D}_{guided}$, $y_i$ stems from questions and masked trajectories. In $\mathcal{D}_{explore}$, $y_i$ stems from questions alone. We maintain a 7:3 ratio between these subsets. This balances structured learning with exploration.

\section{Large-scale Agent Training}
We adopt a two-stage training paradigm, consisting of supervised fine-tuning (SFT) followed by reinforcement learning (RL) optimization.
\subsection{Stage 1: Supervised Fine-Tuning}
According to the KISA method described in section \ref{sec:kisa}, we now have the training dataset $\mathcal{D}_{train}=\{(x_i, y_i)\}_{i=1}^N$. $x_i$ are the medical questions. $y_i$ are correct answers and real reasoning trajectories with tool invocations. Specifically, $y_i$ is serialized with a sequence of tokens.
\begin{equation*}
    y_i=\{y_{i,1}, y_{i,2}, ..., y_{i,T_i}\}
\end{equation*}
Each token $y_{i,t}$ might be an agent token (plain text or tool-call tokens) or a tool output token. Then the objective of the supervised fine-tuning is 
\begin{equation*}
    L_{\text{SFT}}(\theta)=-\frac{1}{N}\sum_{i=1}^{N}\sum_{t=1}^{T_i}m_{i,t}\log p_{\theta}(y_{i,t}|x_i, y_{i,<t})
\end{equation*}
where $p_\theta(y_{i,t}|x_i, y_{i,<t})$ denotes the probability of the agent parameterized with $\theta$ generating $y_{i,t}$ given the question $x_i$ and the previous tokens $y_{i,<t}$. $m_{i,t}$ is an indicator function that follows
\begin{equation*}
    m_{i,t}=\begin{cases}
    &\hspace{-0.75em}1, \hspace{0.5em} \text{if } y_{i,t}\text{ is an agent token}\\
    &\hspace{-0.75em}0, \hspace{0.5em} \text{if } y_{i,t}\text{ is a tool output token}
    \end{cases}
\end{equation*}

We trained the agent from a base model Qwen2.5-32B-Instruct \cite{qwen2.5, qwen2}  through SFT on 2,100+ synthetic medical trajectories generated by our KISA framework. The training incorporates robustness augmentations including tool failure simulation (5\% corruption rate), intermediate thought supervision, and multi-task sampling across medical domains. This stage establishes fundamental tool usage patterns and medical reasoning capabilities. Detailed training configurations are provided in Appendix~\ref{app:training_setup}.

\subsection{Stage 2: Reinforcement Learning}
Following SFT, we refine the agent using Grouped Regularized Policy Optimization (GRPO) with composite rewards balancing task accuracy, expert alignment, and efficiency. For reinforcement learning training, the agent is modeled as a policy $\pi_\theta$ parameterized by $\theta$. The reward is formulated as:
\begin{equation*}
    r = \alpha r_{\text{task}} + \beta r_{\text{expert}} - \gamma r_{\text{efficiency}}
\end{equation*}
where $r_{\text{task}}$ is the accuracy of the final predictions, $r_{\text{expert}}$ is a strong LLM evaluation on the medical accuracy and completeness, $r_{\text{efficiency}}$ is a penalty for redundant tool usage.
% where $\alpha=1.0$, $\beta=0.2$, $\gamma=0.1$. 
For GRPO, we conduct $G$ rollouts $\{o_1, o_2, ..., o_G\}$ calculate the rewards and advantages $\hat{A}_{i,t}$ within the rollouts. Then the objective becomes
\begin{equation*}
\resizebox{0.8\columnwidth}{!}{$
\begin{aligned}
    \mathcal{J}=&\frac{1}{G}\sum_{i=1}^{G}\frac{1}{|o_i|}\sum_{t=1}^{|o_i|}\min\{\frac{\pi_{\theta}(o_{i,t}|q, o_{i,<t})}{\pi_{\theta_{ref}}(o_{i,t}|q, o_{i,<t})}\hat{A}_{i,t},
    \text{clip}\left(\frac{\pi_{\theta}(o_{i,t}|q, o_{i,<t})}{\pi_{\theta_{ref}}(o_{i,t}|q, o_{i,<t})},1-\epsilon,1+\epsilon\right)\hat{A}_{i,t}\}
\end{aligned}
$}
\end{equation*}

We deliberately omit KL regularization following recent findings~\citep{he2025skywork} and implement curriculum learning for progressive task complexity. The GRPO objective employs group-level baseline normalization for stable gradient estimation. Other details are shown in Appendix~\ref{app:training_setup}.

% \vspace{-0.3em}
\section{Experiments}

% We evaluate our MedResearcher-R1 agent on three medical domain benchmarks. The results demonstrate that MedResearcher-R1 exhibits strong performance in medical deep research problems.

% We evaluate {MedResearcher-R1} across both a specialized medical benchmark and two general-purpose agent benchmarks. This dual evaluation framework is designed to rigorously assess its primary capability in complex medical research while simultaneously measuring its generalization ability in open-domain tasks.

% \vspace{-0.4em}
\subsection{Benchmarks}
We evaluate MedResearcher-R1 using three public benchmarks: MedBrowseComp, GAIA, and XBench-DeepSearch. While MedBrowseComp focuses on the medical domain, the latter two serve as general-purpose benchmarks, allowing us to assess MedResearcher-R1's performance across both specialized medical tasks and general inquiries.

\textbf{MedBrowseComp}~\citep{chen2025medbrowsecompbenchmarkingmedicaldeep} is designed to assess the capabilities of LLM-based agents in answering medical questions that require multi-hop reasoning. This benchmark comprises 50 questions drawn from a wide range of medical topics. Each question requires a detailed exploration of medical databases, literature, and clinical knowledge. Agents are evaluated based on their accuracy on these 50 questions.

\textbf{GAIA}~\citep{mialon2023gaiabenchmarkgeneralai} is a general-purpose benchmark that evaluates assistant capabilities in tasks requiring multi-modal tool use, web search, and multi-step reasoning. We utilize a subset of 103 cases from the text-only validation set to test the agent’s ability to understand complex scenarios, reason effectively, and interact with tools to generate responses.

\textbf{XBench-DeepSearch}~\citep{chen2025xbenchtrackingagentsproductivity} is a multi-domain benchmark focused on evaluating tool usage capabilities in search and information retrieval scenarios. It tests the agent’s ability to perform advanced information synthesis across various domains.

% \vspace{-0.3em}
% \subsection{Baselines}
% \vspace{-0.2em}
% To provide a comprehensive comparison, we evaluate {MedResearcher-R1} against several deep research agents. 

% Our model, {MedResearcher-R1}-32B, is built upon the Qwen-2.5-32B model, making it the most direct baseline for assessing the impact of our methods.
% \begin{itemize}[leftmargin=15pt, itemsep=-0.15em]
% \vspace{-0.5em}
%     \item \textbf{Medical Domain Baselines}: We compare against leading proprietary deep research systems known for their strong reasoning capabilities. These include Google's Gemini-2.5-Pro deepsearch and OpenAI's o3-deepresearch, which represent the state-of-the-art in closed-source deep research agents.
%     \item \textbf{General Domain Baselines}: For open-domain tasks, we compare against a range of top-performing models. This includes powerful proprietary models like GPT-4o and o4-mini, as well as leading open-source deep research agent WebSailor~\citep{li2025websailor}, which is the current state-of-the-art open-source agent for general deep research tasks.
% \end{itemize}

\subsection{Main Results}
\textbf{State-of-the-Art Performance in Medical Research.}
As shown in Table~\ref{tab:med-results}, {MedResearcher-R1} establishes a new state-of-the-art on the challenging MedBrowseComp benchmark. With a score of \textbf{27.5/50}, it outperforms strong proprietary systems like o3-deepresearch (25.5) and Gemini-2.5-Pro-deepsearch (24.5). This result validates the effectiveness of our specialized data synthesis and domain-specific tools in equipping the agent with superior medical reasoning capabilities.
% \begin{table}[h]
% \centering
% \resizebox{\linewidth}{!}{
%     \begin{tabular}{|l|c|}
%     \toprule
%     &MedBrowseComp (Accuracy $\uparrow$) \\
%     \midrule
%     o3 search & 19.0/50\\
%     Gemini-2.5-pro Deepsearch & 24.5/50 \\
%     o3 deepsearch & 25.5/50\\
%     claude-cua & 18.0/50\\
%     MedResearcher-R1 (Ours) & \textbf{27.5/50}\\
%     \bottomrule
%     \end{tabular}}
% \caption{Performance Comparison on MedBrowseComp Benchmarks (number correct out of 50)}
% \label{tab:med-results}
% \vspace{-1em}
% \end{table}
\begin{table}[t]
\centering
\resizebox{0.6\linewidth}{!}{
    \begin{tabular}{l|c}
    \toprule
    &\textbf{MedBrowseComp} (Accuracy $\uparrow$) \\
    \midrule
    o3 search & 19.0/50\\
    Gemini-2.5-pro Deepsearch & 24.5/50 \\
    o3 deepsearch & 25.5/50\\
    claude-cua & 18.0/50\\
    MedResearcher-R1 (Ours) & \textbf{27.5/50}\\
    \bottomrule
    \end{tabular}}
\caption{Performance Comparison on MedBrowseComp Benchmarks (number correct out of 50).}
\label{tab:med-results}
\end{table}

\noindent
\textbf{Strong Generalization to Open-Domain Tasks.}
Notably, our specialization in the medical domain does not come at the cost of general capabilities. Table~\ref{tab:main-results} shows that {MedResearcher-R1} achieves highly competitive performance on general agent benchmarks. On both GAIA (53.4) and XBench-DeepSearch (54.0), our 32B model performs on par with the open-source agent, WebSailor-32B (53.2 and 53.3, respectively). This demonstrates that the complex reasoning patterns and robust tool-use strategies learned from the medical domain transfer effectively to general problem-solving scenarios.

\begin{table}[h]
\centering
\resizebox{0.7\linewidth}{!}{%
\begin{tabular}{lccc}
\toprule
\textbf{Model} & \textbf{Paradigm} & \textbf{Xbench-DeepSearch} & \textbf{GAIA} \\
\midrule
Qwen-2.5-32B & Direct & 8.7 & 13.6 \\
Qwen-2.5-72B & Direct & 12.7 & 14.6 \\
GPT-4o & Direct & 18.0 & 17.5 \\
GPT-4.1 & Direct & 17.0 & 22.3 \\
QwQ-32B & Direct & 10.7 & 22.3 \\
o4-mini & Direct & 22.3 & 33.3 \\
DeepSeek-R1 & Direct & 32.7 & 16.5 \\
Qwen-2.5-32B & Search-o1 & 3.7 & 28.2 \\
WebDancer-32B & ReAct & 38.7 & 40.7 \\
QwQ-32B & Search-o1 & 25.0 & 39.8 \\
WebSailor-7B & ReAct & 34.3 & 37.9 \\
WebSailor-32B & ReAct & 53.3 & 53.2 \\
MedResearch-R1-32B(Ours) & ReAct & \textbf{54.0} & \textbf{53.4} \\
\bottomrule
\end{tabular}%
}
\caption{Performance Comparison on Xbench-DeepSearch and GAIA Benchmarks}
\label{tab:main-results}
\end{table}

\subsection{Qualitative Analysis}
% \vspace{-0.2em}
To understand the underlying factors driving performance improvements, we conducted an in-depth analysis of the training data patterns and their impact on agent behavior. For medical deep research problems, the workflow based on an iterative think–tool invocation-verification paradigm might be the most significant improvement in deep research capabilities. Given the dynamic nature of medical knowledge, continuous cross-validation is essential for ensuring factual accuracy and correcting outdated information.

\begin{figure}[ht]
\centering
\includegraphics[width=0.975\linewidth]{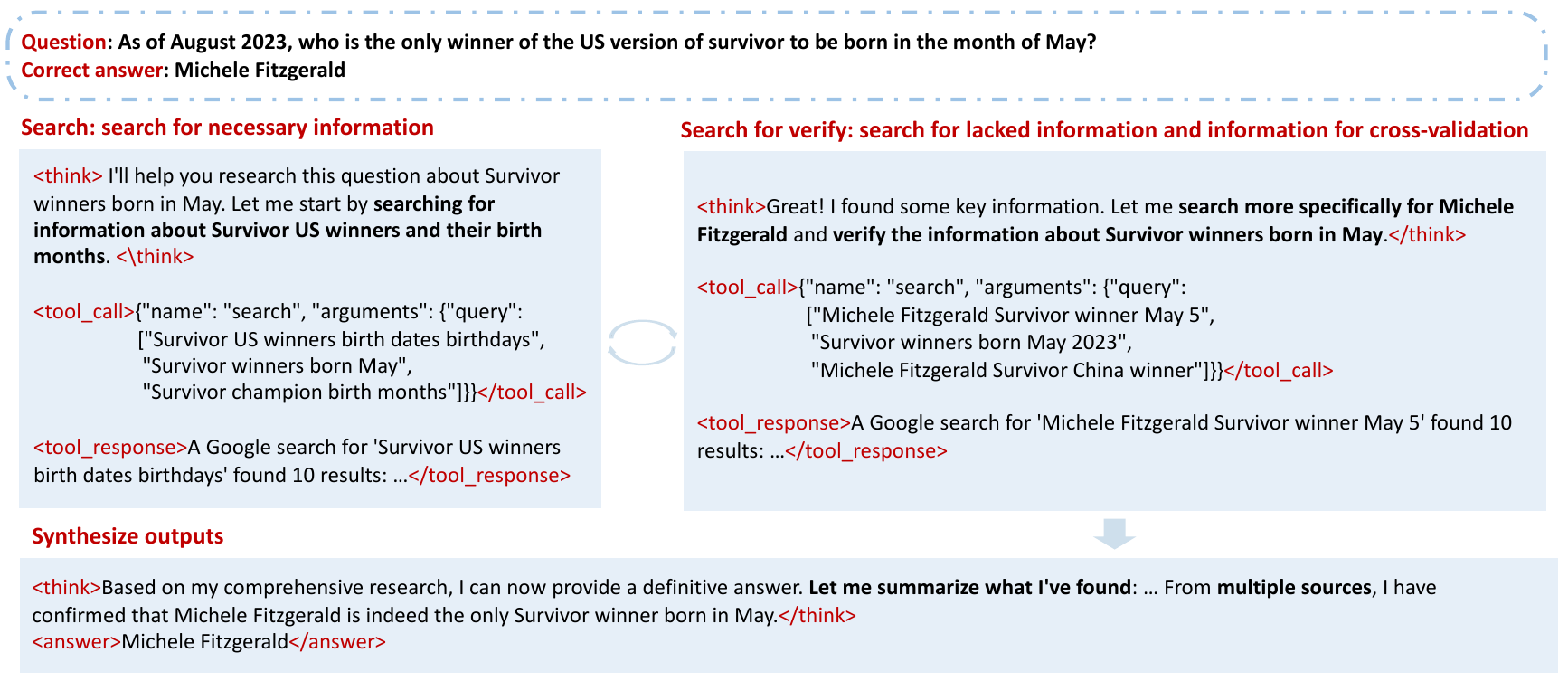}
\caption{\textbf{Sample trajectory demonstrating the thinking-tool invocation-verification workflow}. The image depicts the agent’s workflow for answering a given question. In the “think” phase, the agent performs (i) searching for necessary information, (ii) identifying missing information and selecting evidence that requires cross-validation. Then the verification step searches the required information and continues the loop. Finally the agent summarizes and synthesizes the retrieved information into an output.}
\label{fig:case-study}
% \vspace{-0.5em}
\end{figure}

As illustrated in Figure~\ref{fig:case-study}, our agent achieves superior research depth through systematic evidence synthesis. The methodical approach relies on multiple cross-validation cycles to ensure response consistency. This make it contrast sharply with baseline models, which often exhibit premature convergence or suboptimal tool-use patterns.

\textbf{Cross validations from different resources are significant for medical deep research questions}. We performed a comparative analysis of successful trajectories, contrasting our cross-validation pipeline against a self-verification baseline. In cross-validation, the agent searches for the same information from different sources and draws a conclusion based on an analysis over all information. As comparison, a self-validation agent is set to self-check about whether the retrieved information is valid for the problem to be solved. Our results indicate that trajectories employing cross-validation achieve a 34.2\% higher success rate on complex multi-hop reasoning tasks. This iterative approach ensures factual grounding and response convergence, which are vital for high-precision domains like medical diagnostics. These findings suggest that the efficacy of tool-augmented agent training is tied to the structural complexity of the training data, with iterative cross-validation serving as a driver of robust research capabilities.

\section{Ablation Study}
\label{sec:ablation}
To isolate the factors driving performance gains, we systematically remove some components from MedResearcher-R1 to isolate their contributions. The ablation results are shown in Table \ref{tab:ablation_main}. From the results, we can draw following conclusions.

\begin{table}[t]
\centering
\resizebox{0.95\linewidth}{!}{%
\begin{tabular}{@{}lcccc@{}}
\toprule
\textbf{Model Configuration} & \textbf{MedBrowseComp (correct/50)} & \textbf{GAIA (\%)} & \textbf{XBench-DeepSearch (\%)} & \textbf{Avg. Tool Calls} \\
 % & \textbf{(correct/50)} & \textbf{(\%)} & \textbf{(\%)} &  \\
\midrule
MedResearcher-R1 (Full) & \textbf{27.5} & \textbf{53.4} & \textbf{54.0} & 4.2 \\
\midrule
\multicolumn{5}{@{}l}{\textit{Component Ablations}} \\
\quad w/o Medical Tools          & 23.1        & 48.3    & 40.0    & 3.3 \\
\quad w/o MTG                    & 24.2  & 44.3 & 47.8 & 3.5 \\
% \quad w/o Rare Entities          & 20.1$^*$  & 27.8$^*$ & 38.2$^*$ & 3.2 \\
\midrule
\multicolumn{5}{@{}l}{\textit{Data Ablations}} \\
\quad Common Entities Only       & 23.0  & 43.0 & 46.0 & 4.5 \\
% \quad No Tool Diversity          & 21.0$^*$  & 38.0$^*$ & 49.0      & 3.2 \\
\midrule
\multicolumn{5}{@{}l}{\textit{Training Ablations}} \\
\quad SFT Only                   & 25.5  & 49.0     & 48.0 & 3.4 \\
\quad RL Only (no SFT)           & 12.0  & 34.0 & 34.0 & 3.2 \\
\bottomrule
\end{tabular}%
}
\caption{\textbf{Ablation study for MedResearcher-R1.} Remove key components while keeping all other settings fixed. Component ablations include: removing the medical tool sets, and generating reasoning trajectories without the masked reasoning path. Data ablations consist of constructing questions using only common entities. Training ablations include using SFT only and using RL only without SFT training.}
\label{tab:ablation_main}
\end{table}

\noindent
\textbf{Tool sets and MTG method are important for agent}. The integration of medical tools markedly enhances performance ($40.0$ to $54.0$ on XBench), underscoring that specialized tool environments are essential for complex reasoning. MTG also improve the agent performance. Our analysis of the MTG-derived SFT data reveals a fundamental requirement for training dataset design: agents benefit most from heterogeneity that balances the exploitation of long reasoning chains with exploratory trajectories (our data mix of $D_{guided}$ and $D_{explore}$). This dual approach is key to developing agents capable of dealing with complex medical research.

\noindent
\textbf{Rare entities-based KISA are crucial.} Restricting the knowledge graph to common entities results in a significant performance degradation, with scores on MedBrowseComp falling from $27.5$ to $23.0$. This decline suggests that common-entity graphs primarily generate queries with low reasoning complexity and high information accessibility. Consequently, models trained on such datasets fail to develop the sophisticated ability required for multi-hop reasoning. These results underscore that agent proficiency in solving complex problems is contingent upon the structural complexity and informational density of the knowledge graph.

\noindent
\textbf{Supervised Fine-tuning works better when given high-quality training data.} SFT alone achieves $25.5/50$ on MedBrowseComp while RL alone fails ($12.0/50$). The relative failure of RL can be attributed to the sparse reward signals inherent in long-horizon medical reasoning. Without the structural priors provided by high-quality SFT trajectories, the agent's search space is too vast to explore effectively. This suggests that explicit behavioral guidance on expert-level reasoning chains is a prerequisite for stabilizing agentic performance before RL optimization.

\section{Conclusion}
In this work, we introduce MedResearcher-R1. It is a specialized medical research agent fine-tuned on data synthesized via our KISA framework. KISA addresses the scarcity of high-quality SFT data by systematically generating challenging, multi-hop medical queries paired with dense reasoning trajectories. This ensures that agents are exposed to the compositional complexity inherent in clinical research. MedResearcher-R1 integrating an iterative "think–tool invocations–verification" paradigm with a suite of domain-specific tools. The SFT with training data plus the special work paradigm make it achieve a state-of-the-art (SOTA) pass@1 accuracy of on MedBrowseComp, while maintaining robust generalization across broader agent benchmarks such as GAIA and XBench-DeepSearch.

Our empirical evaluation shows that training on KISA-generated data improves agent performance. Additionally, the integration of specialized medical tools and our custom workflow contributes to the high performance observed in the medical deep research domain. We believe that KISA-generated training data for agent SFT can help agents achieve more stable and better training results.

% \newpage
% \vspace{-0.3em}
\section*{Limitations}
% \vspace{-0.4em}
While our knowledge graph (KG)-based framework effectively synthesizes training data for complex research tasks, its generalizability to broader agentic domains remains to be fully explored. The current methodology is optimized for the structural properties of deep research problems, specifically those requiring multi-step reasoning and factual tracing. This may not be directly applicable to task-oriented agents (e.g., UI automation or creative generation). Furthermore, we acknowledge that medical research is increasingly multi-modal. However, our framework is not capable of generating training data for multi-modal deep search problems. %Therefore, we plan to extend our framework to support multi-modal deep research. We intend to develop an augmented knowledge graph capable of capturing cross-modal dependencies. This can enable agent to synthesize evidence from both unstructured text and diverse medical imaging modalities.

% \newpage

\bibliographystyle{plainnat}
\bibliography{custom}

\appendix
\section*{Appendix}
\section{Training setting}
\label{app:training_setup}

\subsection{Supervised Fine-Tuning Configuration}
\label{app:sft_details}

\paragraph{Dataset.} The final training dataset ($\mathcal{D} = \{(x^{(i)}, y^{(i)})\}_{i=1}^N$) generated by KISA contains a total 2,137 questions with corresponding trajectories ($N=2,137$). The objective function is:
\begin{equation}
\resizebox{0.5\columnwidth}{!}{$
\begin{aligned}
L_{\text{SFT}}(\theta)=-\frac{1}{N}\sum_{i=1}^{N}\sum_{t=1}^{T_i}m_{i,t}\log p_{\theta}(y_{i,t}|x_i, y_{i,<t})
\end{aligned}
$}
\end{equation}
where $p_\theta(y_{i,t}|x_i, y_{i,<t})$ denotes the probability of the agent parameterized with $\theta$ generating $y_{i,t}$ given the question $x_i$ and the previous tokens $y_{i,<t}$. $T_i$ is the token length of the $i$-th sample trajectory. $m_{i,t}$ is an indicator function that follows
\begin{equation*}
    m_{i,t}=\begin{cases}
    &\hspace{-0.75em}1, \hspace{0.5em} \text{if } y_{i,t}\text{ is an agent token}\\
    &\hspace{-0.75em}0, \hspace{0.5em} \text{if } y_{i,t}\text{ is a tool output token}
    \end{cases}
\end{equation*}

\noindent
\textbf{Robustness Augmentations.}
\begin{itemize}[leftmargin=10pt, itemsep=-0.2em]
\item \textbf{Tool failure simulation:} We introduce a 5\% random corruption rate of tool outputs to encourage error recovery. Specifically, there is a 5\% of probability that an API return will be replaced with a blank response. This simulates scenarios where information retrieval fails during the multi-turn thinking process. Therefore, this enhances the agents' capacity to re-search the information.
% \item \textbf{Intermediate thought supervision:} Explicit reasoning traces before each tool invocation
\item \textbf{Multi-task sampling:} Balanced batching across different medical question types. For example, diagnosis (30\%), treatment (25\%), guidelines (25\%), rare diseases (20\%).
\end{itemize}

\noindent
\textbf{Optimization.}
\begin{itemize}[leftmargin=10pt, itemsep=-0.2em]
\item Optimizer: AdamW with $\beta_1=0.9$, $\beta_2=0.98$
\item Learning rate: $\lambda=0.01$ with cosine annealing to $\eta_{\min}=3\times10^{-7}$
\item Batch size: 128 (16 per GPU $\times$ 8 H800 GPUs)
\item Training epochs: 3
\item Gradient clipping: 1.0
\item Warmup steps: 100
\end{itemize}

\subsection{Reinforcement Learning Configuration}
\label{app:rl_details}

\noindent
\textbf{Reward Components.}
The composite reward function $r = \alpha r_{\text{task}} + \beta r_{\text{expert}} - \gamma r_{\text{efficiency}}$ comprises:

\begin{itemize}[leftmargin=10pt, itemsep=-0.2em]
\item $r_{\text{task}}$: Binary task completion (1.0 for correct, 0.0 for incorrect)
\item $r_{\text{expert}}$: GPT-4 preference score $\in [0,1]$ evaluating medical accuracy and completeness
\item $r_{\text{efficiency}}$: Penalty for redundant tool usage, computed as:
\begin{equation}
\resizebox{0.75\columnwidth}{!}{$
\begin{aligned}
r_{\text{efficiency}} =& 0.1 \times n_{\text{redundant}} +
 0.2 \times n_{\text{post-answer}} + 0.15 \times n_{\text{irrelevant}}
\end{aligned}
$}
\end{equation}
\end{itemize}
where $n_{\cdot}$ is the tool usage count of the footnote type.

\paragraph{GRPO Configuration.}
For GRPO, we conduct $G$ rollouts $\{o_1, o_2, ..., o_G\}$ calculate the rewards and advantages $\hat{A}_{i,t}$ within the rollouts. Then the objective for reinforcement learning is: 
\begin{equation*}
\resizebox{0.8\columnwidth}{!}{$
\begin{aligned}
    \mathcal{J}=&\frac{1}{G}\sum_{i=1}^{G}\frac{1}{|o_i|}\sum_{t=1}^{|o_i|}\min\{\frac{\pi_{\theta}(o_{i,t}|q, o_{i,<t})}{\pi_{\theta_{ref}}(o_{i,t}|q, o_{i,<t})}\hat{A}_{i,t},
    \text{clip}\left(\frac{\pi_{\theta}(o_{i,t}|q, o_{i,<t})}{\pi_{\theta_{ref}}(o_{i,t}|q, o_{i,<t})},1-\epsilon,1+\epsilon\right)\hat{A}_{i,t}\}
\end{aligned}
$}
\end{equation*}

Detailed settings are:

\begin{itemize}[leftmargin=10pt, itemsep=-0.2em]
\item Group size: 4 responses per query
\item Sampling temperature: 0.7
\item PPO clip range: 0.2
\item Value loss coefficient: 0.5
\item Entropy coefficient: 0.01
\item Training iterations: 500
\item KL regularization: Disabled (following \citet{he2025skywork})
\end{itemize}

% \textbf{Curriculum Learning.}
% Task complexity increases based on moving average pass rate:
% \begin{itemize}[leftmargin=25pt, itemsep=-0.2em]
% \item Level 1 (pass rate $>$ 80\%): Single-hop queries
% \item Level 2 (pass rate $>$ 60\%): 2-3 hop queries  
% \item Level 3 (pass rate $>$ 40\%): 4+ hop queries with rare entities
% \end{itemize}

\section{Prompt}
Here we show the prompts for constructing reasoning trajectories with masked trajectory guidance. The system prompts are:
\begin{lstlisting}
role_definition = "You are a professional information research assistant"
        base_content = '''. Your core responsibility is to solve complex questions through intelligent tool usage and logical reasoning, providing accurate and reliable answers.

**Working Methodology**:

1. **Problem Analysis**: First analyze the nature of the question and required information types:
   - **Basic Reasoning**: Mathematical calculations, logical deduction, common sense judgments should be completed directly
   - **External Information**: Only use tools when you need real-time data, specific facts, or web content

2. **Intelligent Tool Usage**: Use tools only when external information is needed:
   - Search for unknown facts, data, news, etc.
   - Retrieve webpage content and specific information
   - Verify time-sensitive information

3. **Reasoning First**: For content that can be deduced through reasoning, analyze directly:
   - Mathematical operations and logical judgments
   - Time calculations and numerical relationships
   - Inference based on known information

4. **Comprehensive Analysis**: Combine external information obtained through tools with autonomous reasoning to provide complete answers.

**Working Principles**: Wisely choose when to use tools and when to rely on autonomous reasoning, ensuring efficient and accurate problem-solving.'''
\end{lstlisting}

The main prompts are:
\begin{lstlisting}
"""Do research on the question and answer it when you finish the research. When you finish your research, you should explain first and then answer, your answer should be place inside <answer></answer>, and your answer should be direct answer without any explanation. 

**IMPORTANT**: You are provided with some search guidance hints below. These hints suggest potential SEARCH DIRECTIONS and INVESTIGATION APPROACHES. You MUST NOT use any specific information from these hints directly as your answer or for direct reasoning.

## Search Guidance (Use ONLY as search directions):
{reasoning_path}

## Critical Instructions:

1. **Search Direction Only**: The above hints are ONLY suggestions for what topics/keywords to search for and what aspects to investigate. They are NOT factual information to be used directly in your reasoning or answers.

2. **Mandatory Tool Verification**: For any information mentioned in the guidance hints, you MUST use tools to independently find, verify, and confirm that information. Never assume the hints contain accurate facts.

3. **Balanced Approach**: 
   - **Basic reasoning** (mathematical calculations, logical deductions, common sense): Handle directly without tools
   - **Information from hints** (specific facts, data, claims): MUST be verified through tools
   - **External information** (current events, specific details): Use tools to search and verify

4. **Independent Research**: Treat the guidance as a research roadmap only. You must independently discover, verify, and validate all specific information through your tools.

5. **Evidence-Based Answers**: Your final answer must be based on:
   - Verified information you actually found through tools (not from hints)
   - Sound reasoning and calculations you performed directly

**Remember**: The guidance hints may contain inaccurate or incomplete information. Always verify through tools before using any specific claims from the hints.

When you finish your research, explain your findings first and then provide your answer inside <answer></answer>. Your answer should be a direct answer without any explanation inside the answer tags.

User Question: """,
\end{lstlisting}

\end{document}